\pdfoutput=1

\documentclass[11pt]{article}

\usepackage{amsmath,amsfonts,bm}

\def\eqref#1{equation~\ref{#1}}

\def\1{\bm{1}}

\DeclareMathAlphabet{\mathsfit}{\encodingdefault}{\sfdefault}{m}{sl}
\SetMathAlphabet{\mathsfit}{bold}{\encodingdefault}{\sfdefault}{bx}{n}

\usepackage{color}
\usepackage{times}
\usepackage{epsfig}
\usepackage{graphicx}
\usepackage{amsmath}
\usepackage{amsthm}
\usepackage{amssymb}
\usepackage{mathtools}
\usepackage{multirow}
\usepackage{bbm}
\usepackage{dsfont}

\usepackage[table, dvipsnames]{xcolor}

\usepackage[utf8]{inputenc}
\usepackage{booktabs,siunitx}
\usepackage{multirow}
\usepackage{multicol}
\usepackage{amsmath}
\usepackage{subcaption}
\usepackage{caption}
\usepackage{graphicx}
\usepackage{pifont}

\usepackage{tikz}
\usetikzlibrary{shapes,arrows,patterns}
\usetikzlibrary{positioning}
\usetikzlibrary{calc}

\newcommand{\cut}[1]{}

\newcommand{\postspace}{\vskip -3mm}
\newcommand{\minipostspace}{\vskip -2mm}

\definecolor{red}{RGB}{255, 117, 115}
\definecolor{green}{RGB}{171, 255, 175}

\usepackage[final]{acl}

\usepackage{times}
\usepackage{latexsym}

\usepackage[T1]{fontenc}

\usepackage[utf8]{inputenc}

\usepackage{microtype}

\usepackage{inconsolata}

\usepackage{graphicx}

\definecolor{cadmiumgreen}{rgb}{0.0, 0.42, 0.24}
\definecolor{cardinal}{rgb}{0.77, 0.12, 0.23}
\definecolor{cadmiumred}{rgb}{0.89, 0.0, 0.13}

\usepackage[most]{tcolorbox}
\newtcolorbox[list inside=prompt,auto counter,number within=section]{prompt}[1][]{
    fontupper=\ttfamily\footnotesize,
    boxsep=5pt,
    left=0pt,
    right=0pt,
    top=0pt,
    bottom=0pt,
    boxrule=1pt,
    breakable,
    #1,
}

\title{Insight-RAG: Enhancing LLMs with Insight-Driven Augmentation}

\author{Pouya Pezeshkpour\\
Megagon Labs\\
\texttt{pouya@megagon.ai} \\
\And
Estevam Hruschka\\
Megagon Labs \\
\texttt{estevam@megagon.ai}}

\begin{document}
\maketitle
\begin{abstract}
Retrieval Augmented Generation (RAG) frameworks have shown significant promise in leveraging external knowledge to enhance the performance of large language models (LLMs). However, conventional RAG methods often retrieve documents based solely on surface-level relevance, leading to many issues: they may overlook deeply buried information within individual documents, miss relevant insights spanning multiple sources, and are not well-suited for tasks beyond traditional question answering. 
In this paper, we propose \textit{Insight-RAG}, a novel framework designed to address these issues. 
In the initial stage of Insight-RAG, instead of using traditional retrieval methods, we employ an LLM to analyze the input query and task, extracting the underlying informational requirements. In the subsequent stage, a specialized LLM---trained on the document database---is queried to mine content that directly addresses these identified insights.
Finally, by integrating the original query with the retrieved insights, similar to conventional RAG approaches, we employ a final LLM to generate a contextually enriched and accurate response.
Using two scientific paper datasets, we created evaluation benchmarks targeting each of the mentioned issues and assessed Insight-RAG against traditional RAG pipeline. 
Our results demonstrate that the Insight-RAG pipeline successfully addresses these challenges, outperforming existing methods by a significant margin in most cases. These findings suggest that integrating insight-driven retrieval within the RAG framework not only enhances performance but also broadens the applicability of RAG to tasks beyond conventional question answering. We released our dataset and code\footnote{\url{https://github.com/megagonlabs/Insight-RAG}}.
\end{abstract}

\section{Introduction}
Recent advancements in large language models (LLMs) have spurred renewed interest in Retrieval Augmented Generation (RAG) frameworks \citep{gao2023retrieval, fan2024survey}. RAG has emerged as a powerful solution for mitigating inherent challenges in LLMs—such as hallucination and the lack of recent information—by integrating external document repositories with retrieval models to produce contextually enriched responses.  
However, conventional RAG pipelines typically rely on surface-level relevance metrics for document retrieval, which can result in several limitations: they may overlook deeply buried information within individual documents and miss relevant insights distributed across multiple sources. 
Beyond these retrieval challenges, traditional RAG frameworks lack well-defined solutions for tasks that extend beyond standard question answering.

\begin{figure}[t!]
    \centering
    \includegraphics[width=\linewidth]{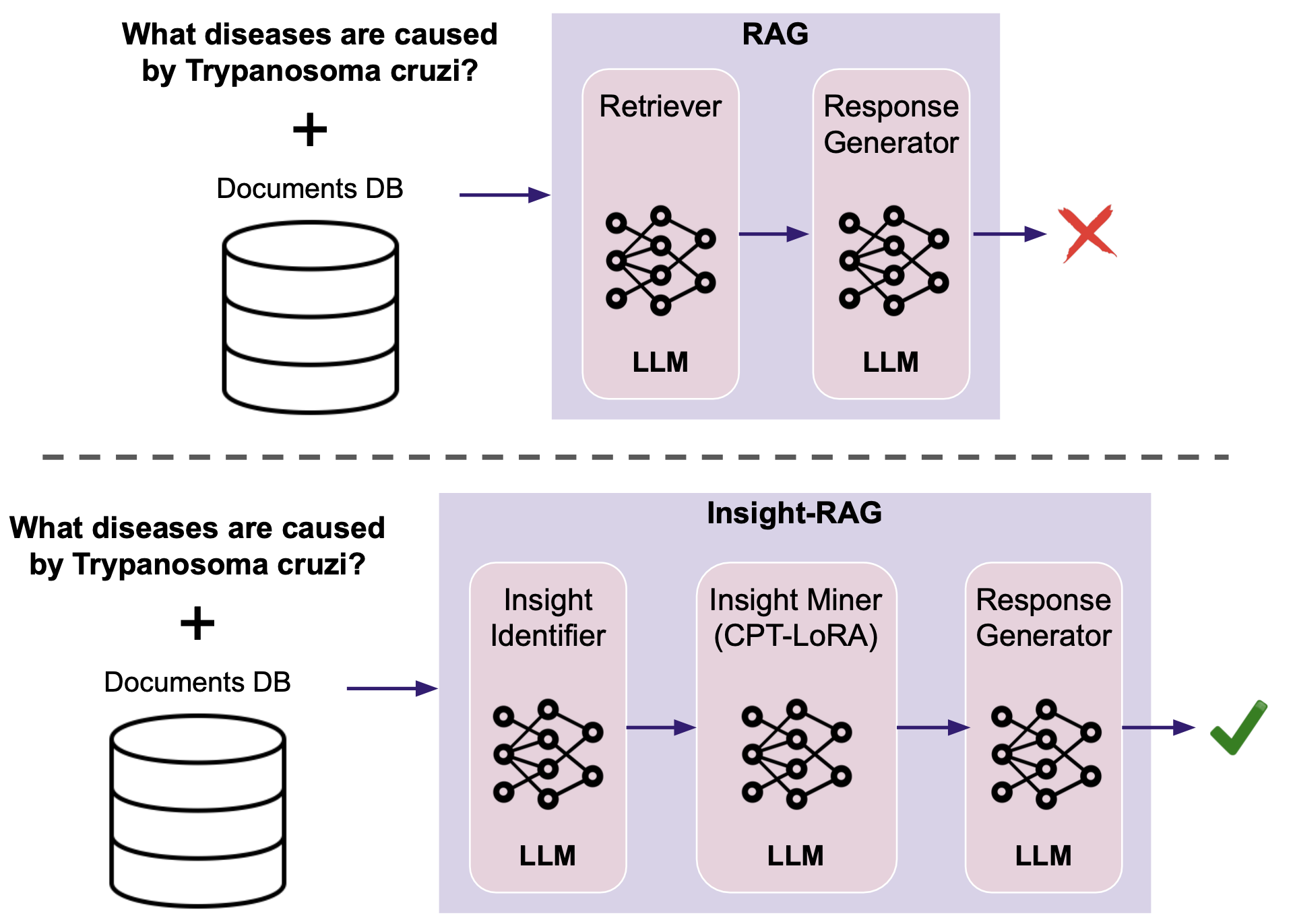}
    \caption{In conventional RAG, using a retriever model, we first retrieve relevant documents to answer a question. In contrast, in \textbf{Insight-RAG}, we first identify necessary insights to solve the task (e.g., answering a question), and then feed the identified insights to an LLM continually pre-trained over the documents to extract the necessary insights before feeding them to the final LLM to solve the task.}
    \label{fig:overview}
    \postspace
\end{figure}

Traditional retrieval mechanisms often fail to capture the nuanced insights required for complex tasks \citep{barnett2024seven, agrawal2024mindful, wang2024astute}. For example, they may overlook deeply buried details within a single document—such as subtle contractual clauses in a legal agreement or hidden trends in a business report—and may neglect relevant insights dispersed across multiple sources, like complementary perspectives from various news articles or customer reviews. Moreover, these methods are not well-equipped for tasks beyond straightforward question answering, such as identifying the best candidate for a job by leveraging insights from a database of resumes or extracting actionable recommendations for business strategy from qualitative feedback gathered from surveys and online reviews.

In this paper, we propose Insight-RAG—a novel framework that refines the retrieval process by incorporating an intermediary insight extraction step (See Figure \ref{fig:overview}). In the first stage, an LLM analyzes the input query and extracts the essential informational requirements, effectively acting as an intelligent filter that isolates critical insights from the query context. This targeted extraction enables the system to focus on deeper, task-specific context. Subsequently, a specialized LLM continually pre-trained \citep{ke2023continual} with LoRA \citep{hu2021lora, zhao2024lora, biderman2024lora} (CPT-LoRA) on the target domain-specific corpus leverages these identified insights to retrieve highly relevant information from the document database. Finally, the original input—now augmented with these carefully retrieved insights—is processed by a final LLM to generate a context-aware response.

To evaluate Insight-RAG, we use two scientific paper datasets—AAN \citep{radev2013acl} and OC \citep{bhagavatula2018content}—and create tailored datasets to address each RAG aforementioned challenge. We sample 5,000 papers from each dataset using a Breadth-First Search strategy and extract triples with GPT-4o mini \citep{hurst2024gpt}, followed by manual/rule-based filtering and normalization. 
For the deeply buried information challenge, we focus on subject-relation pairs that yield a single object, selecting only those triples where both the subject and object appear only once in each document. For the multi-source challenge, we choose subject-relation pairs that yield multiple objects from different documents. 
We then, manually filter the samples after translating each triple into a question using GPT-4o mini. Finally, for the non-QA task challenge, we use the matching labels between papers, capturing the citation recommendation task, provided by \citet{zhou2020multilevel}.

By integrating multiple LLMs to compare Insight-RAG with the conventional RAG approach, we observe that Insight-RAG can achieve up to 60 percentage points improvement in accuracy with much less contextual information, for both deeply buried and multi-source questions. Moreover, we observe that for non-QA tasks such as paper matching, Insight-RAG consistently helps improve performance by up to 5.4 percentage points in accuracy, while using RAG shows mixed results, sometimes increasing and sometimes decreasing the performance. Through various ablation studies, we then connect models behavior to the performance of different components in the pipelines, paving the way for future applications of Insight-RAG.

\begin{figure*}[t!]
    \centering
    \includegraphics[width=0.8\linewidth]{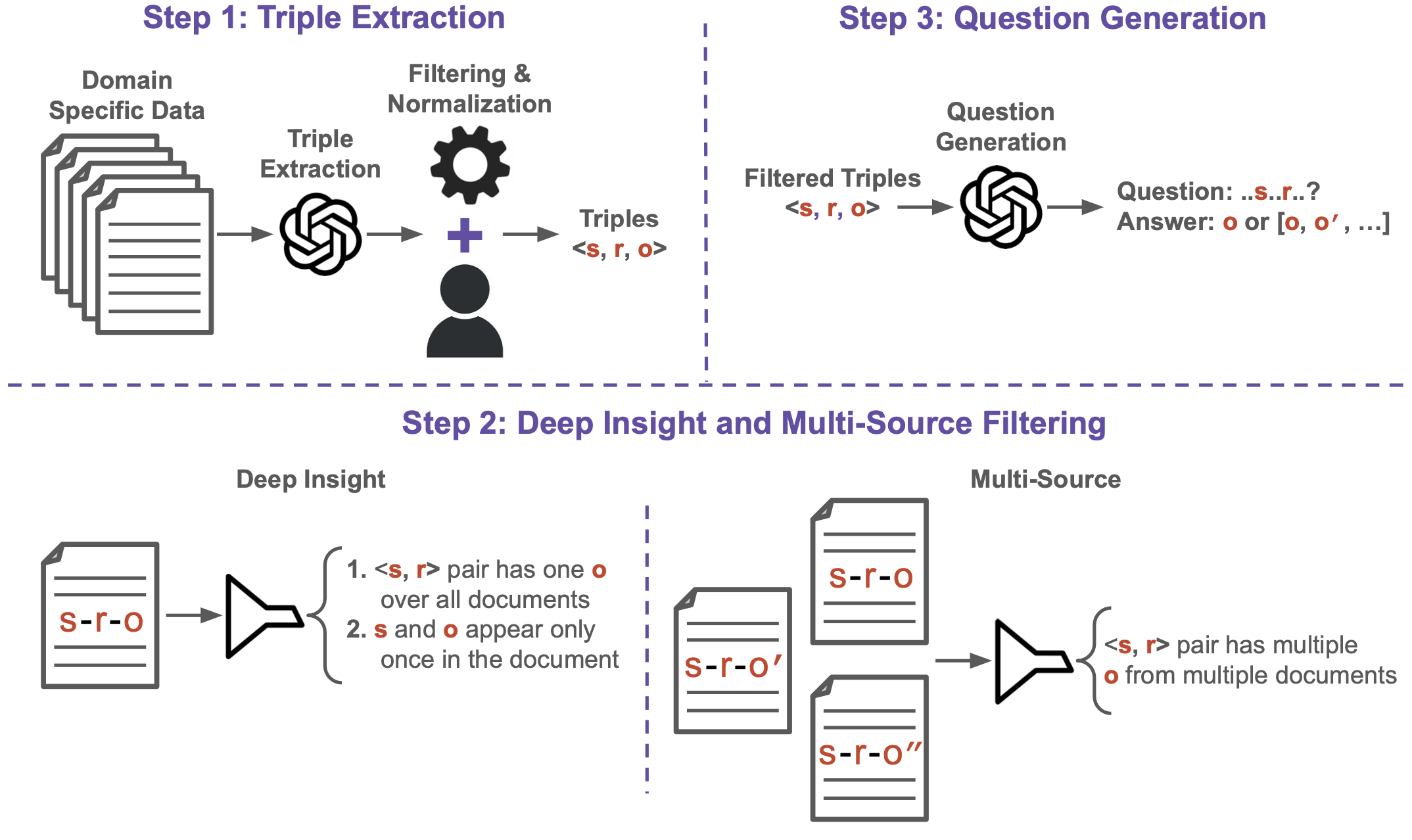}
    \caption{We create our benchmark in several steps: 1) extracting triples from domain-specific documents using GPT-4o mini and then manually normalizing/filtering them, 2) filtering the triples for each different type of issue, 3) using GPT-4o mini to translate the sampled triples to question format, asking about the object of the triple.}
    \label{fig:bench}
    \postspace
\end{figure*}
\section{Insight-RAG}
In this section, we detail our proposed Insight-RAG framework, which consists of three key units designed to overcome the limitations of conventional RAG approaches (see Figure \ref{fig:overview}). By incorporating an intermediary insight extraction stage, our framework captures nuanced, task-specific information that traditional methods often miss. The pipeline comprises the following units:

\paragraph{Insight Identifier:}
The Insight Identifier unit processes the input to extract its essential informational requirements. Serving as an intelligent filter, it isolates critical insights from both the input and the task context, ensuring that subsequent stages concentrate on deeper, necessary content. To facilitate this process, we employ LLMs guided by a carefully designed prompt (provided in the Appendix).

\paragraph{Insight Miner:}
Inspired by previous work \citep{pezeshkpour2025learning}, the insight miner unit leverages a specialized LLM to fetch content for the insights identified earlier. We adopt Llama-3.2 3B \citep{grattafiori2024llama} as our insight-miner, continually pre-training it with LoRA \citep{zhao2024lora, biderman2024lora} over our scientific paper datasets. In line with the previous work on insight mining \citep{pezeshkpour2025learning}, we continually pre-train the model on both the original papers and the extracted triples from them (see Section \ref{sec:bench}). This continual pre-training enables the insight-miner to retrieve highly relevant information to the task.

\paragraph{Response Generator:}
The final unit, response generator, integrates the original query with the retrieved insights and employs a final LLM to generate a comprehensive, context-aware response. Following the conventional RAG approach, this augmented input allows the model to produce outputs that are both accurate and enriched by the additional insights. The prompt used for this stage is provided in the Appendix.


\section{Benchmarking}
\label{sec:bench}
To evaluate the performance of our Insight-RAG framework, we employ two scientific paper's abstract datasets---AAN and OC (provided by \citet{zhou2020multilevel})---to create tailored evaluation benchmarks that address specific challenges encountered in conventional RAG pipelines. Figure \ref{fig:bench} provides an overview of our process for creating the benchmarks. 
Below, we detail our benchmarking process for each identified issue. We provide the data statistics of created benchmarks in Table \ref{tab:stat} and the prompts used in creating the benchmark in the Appendix.

\paragraph{Deeply Buried Insight:}
In this issue, our focus is on the challenge of capturing deeply buried information within individual documents. We begin by sampling 5,000 papers from each dataset using a Breadth-First Search (BFS) strategy. From these papers, following previous works \citep{papaluca2023zero, wadhwa2023revisiting}, we use GPT-4o mini to extract triples (we used the same prompt provided in \citet{pezeshkpour2025learning}), followed by manual/rule-based filtering and normalizing the relations. Then, we select subject-relation pairs that yield a single object and ensure that both the subject and the object appear only once in the paper's abstract. This constraint guarantees that the extracted information is deeply buried and not overly prominent, thereby testing the framework's ability to capture subtle details. We then convert the curated triples into question formats using GPT-4o mini---which generates questions about the object based on the subject-relation pair---and manually filtered them for quality. 
\begin{table}[t!]
\small
\centering
\begin{tabular}{lrr}
\toprule 
& \bf ANN&  \bf OC\\
\midrule
\# Docs&5,000 & 5,000\\
\# Triples&21,526 & 23,662\\
\# Deep-Insight Samples&318 & 403\\ 
\# Multi-Source Samples&173 & 90\\
\# Matching Samples& 500 & 500\\
\bottomrule
\end{tabular}
\caption{Data statistics of the created benchmark.}
\label{tab:stat}
\postspace
\end{table}
\paragraph{Multi-Source Insight:}
To assess the capability of Insight-RAG in synthesizing information from multiple sources, we incorporate the extracted triples from the papers. 
More specifically, we focus on subject-relation pairs that yield multiple objects drawn from different papers, thereby simulating scenarios where relevant insights are distributed across various sources. Once the multi-source triples are curated, we convert them into question formats using GPT-4o mini.  
Acknowledging that some extracted triples may be noisy or vague (e.g., constructs like "<we, show, x>"), we manually filter the questions to ensure quality.

\paragraph{Non-QA Task:}
The third benchmark addresses tasks beyond traditional question answering, specifically evaluating the framework's applicability for citation recommendation. For this benchmark, we leverage the matching labels between papers provided by \citet{zhou2020multilevel}, which capture the citation recommendation task. 
Our goal is to determine if the insights extracted from a document database can effectively support solving arbitrary tasks on inputs that share similarities with the documents, thereby extending the RAG framework’s utility to a variety of real-world applications.

\section{Experimental Details}
We employ several state-of-the-art LLMs as integral components of the Insight-RAG pipeline: GPT-4o \citep{hurst2024gpt}, GPT-4o mini, o3-mini \citep{o3-mini}, Llama3.3 70B \citep{grattafiori2024llama}, and DeepSeek-R1 \citep{guo2025deepseek}. For the Insight Miner unit, we adopt Llama-3.2 3B as our insight-miner, continually pre-trained with LoRA on domain-specific scientific papers and extracted triples. We hyperparameter-tuned the Llama-3.2 3B model based on loss, with additional training and datasets details provided in the Appendix. Moreover, in the Insight-RAG pipeline, we use the same LLM for both the Insight Identifier and Response Generator. 
For RAG Baselines, we used LlamaIndex \citep{Liu_LlamaIndex_2022} and the embedding model gte-Qwen2-7B-instruct \citep{li2023towards}, which is the open-sourced state-of-the-art model based on the MTEB leaderboard \citep{muennighoff2022mteb}. 
Finally, for fair comparison, we limit the insight miner's maximum generated token length to 100 tokens for both datasets, which is less than the average document token length of 134.6 and 226.4 for AAN and OC, respectively. We observe that further increasing the maximum generated token length does not significantly change the performance. 
We evaluate LLM performance using accuracy, exact match accuracy (calculated by determining if the gold response exactly appears in the generated response), and F1 Score (standard QA metrics). We also employ Recall@K, which measures the proportion of correct predictions in the top-k results.


\section{Experiments}
In this section, our goal is to investigate the impact of Insight-RAG in addressing the aforementioned challenges—namely, deeply buried insights, multi-source information, and non-QA tasks. 
We begin by evaluating the considered LLMs using our created benchmarks. By analyzing the models' behavior, we then explore the reasoning behind their performance, examining each component of the Insight-RAG pipeline and assessing the quality of the identified insights.

\begin{figure*}[th!]
    \centering
    \begin{subfigure}[b]{0.49\textwidth}
        \centering
        \includegraphics[width=\linewidth]{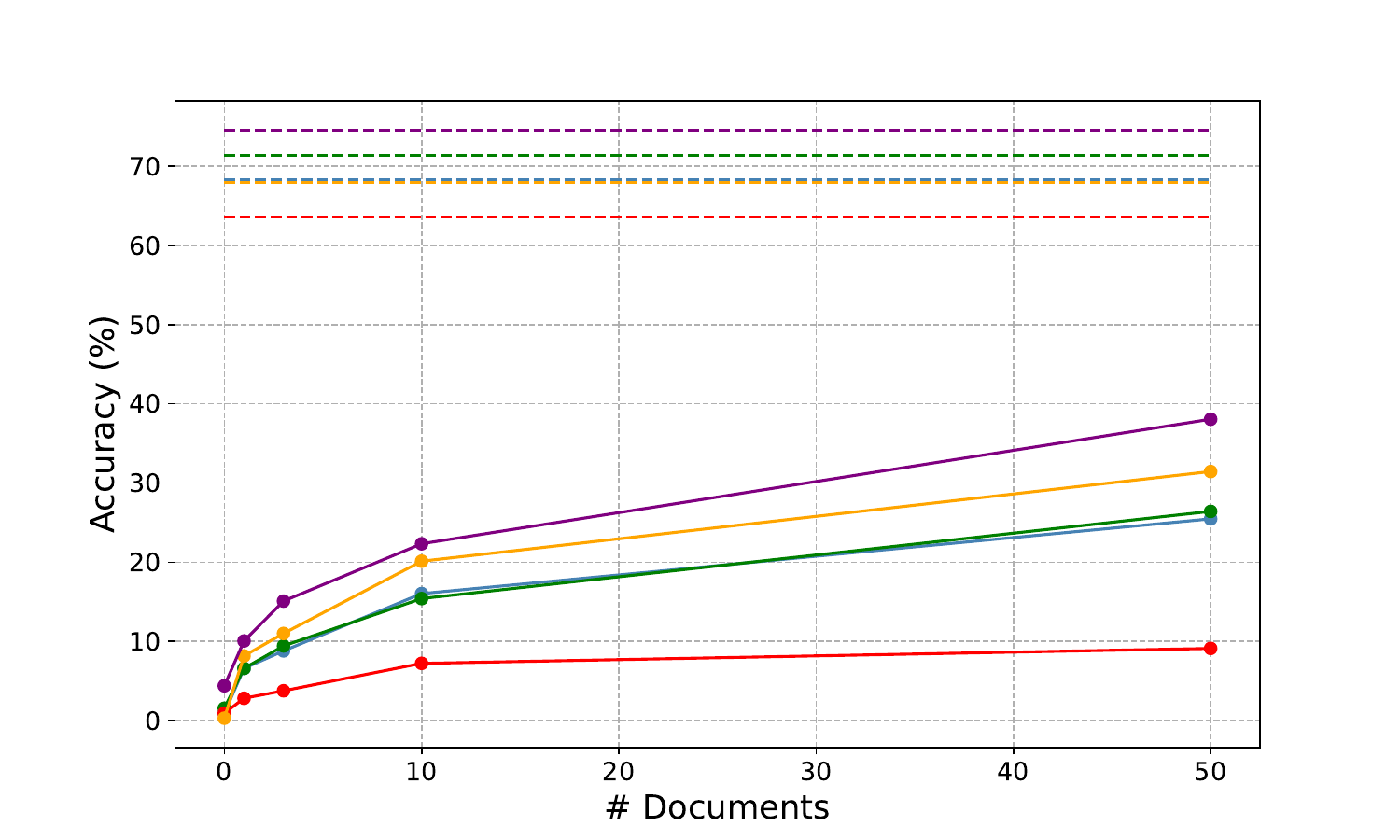}  
        \caption{AAN}
    \end{subfigure}
    \begin{subfigure}[b]{0.49\textwidth}
        \centering
        \includegraphics[width=\linewidth]{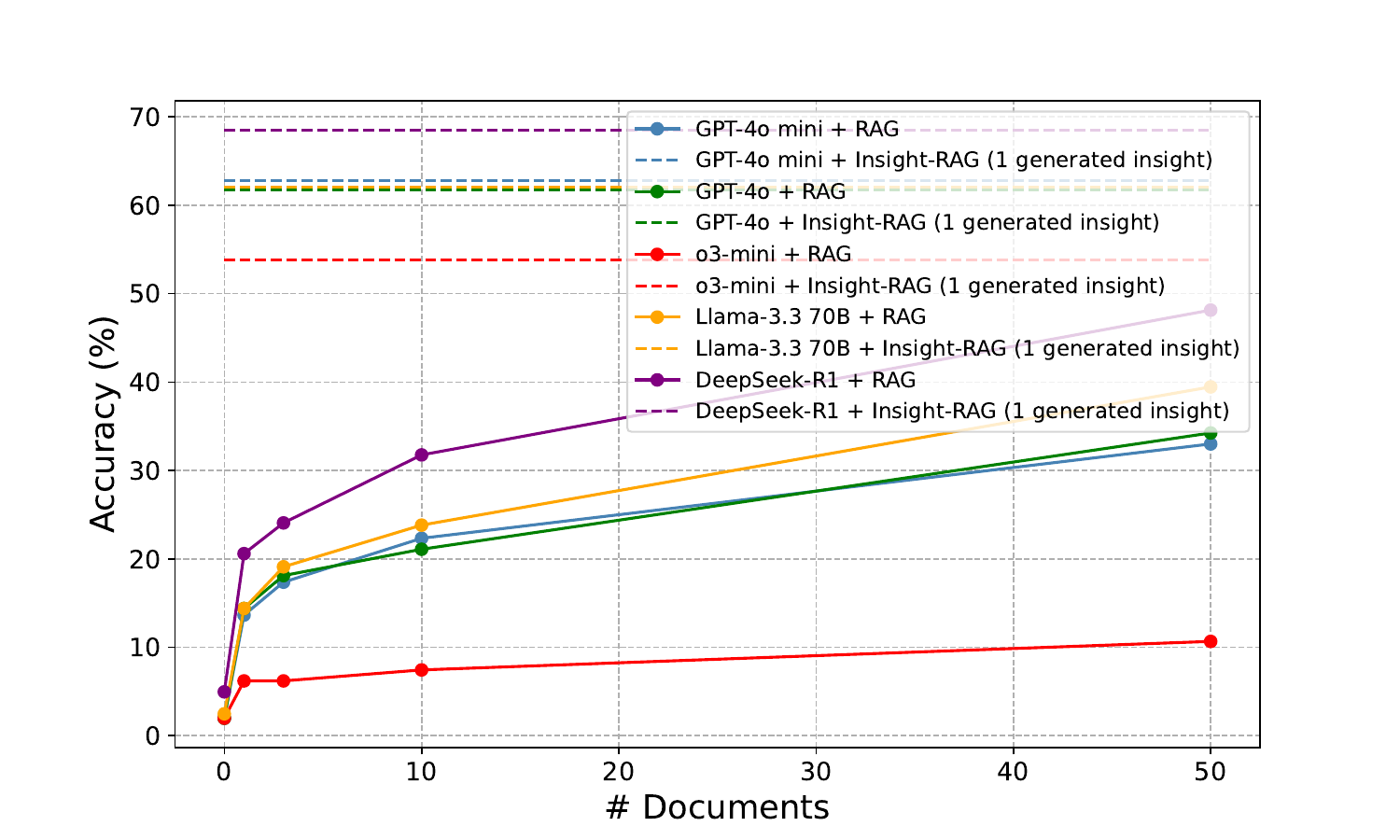}
        \caption{OC}
    \end{subfigure}  
    \caption{The performance comparison of RAG versus Insight-RAG across the AAN and OC datasets in answering question based on deeply buried information. As demonstrated, DeepSeek-R1 performed the best, followed by Llama-3.3 70B. Moreover, we observe that Insight-RAG, even with only one generated insight, outperforms RAG-based solutions by a considerable margin. Additionally, while retrieving more documents reduces this performance gap, Insight-RAG maintains a significant advantage.}
    \label{fig:deep}
    \minipostspace
\end{figure*}
\subsection{Answering Questions using Deeply Buried Insights}
Figure \ref{fig:deep} presents the exact match accuracy of Insight-RAG versus conventional RAG using various LLMs for answering questions based on deeply buried information. First, the zero-shot performance of all LLMs—i.e., without any context or documents—is very low. This is primarily due to the domain-specific nature of the questions, which leaves the LLMs without the necessary information to solve the task. Additionally, the questions themselves may be ambiguous or even erroneous when isolated; however, providing the associated document context alleviates these issues.

As observed, Insight-RAG, even with only one generated insight from the insight miner, achieves significantly higher performance compared to the conventional RAG approach. Although increasing the number of retrieved documents improves the performance of RAG, it still falls considerably short of Insight-RAG. 
We suspect that the shortcomings of the RAG-based solution are due to retrieval errors (as confirmed in Section \ref{sec:comp}) and discrepancies in phrasing between the generated questions and the text, which negatively impact performance \citep{modarressi2025nolima}. 
DeepSeek-R1 performs best, followed by Llama-3.3, both outperforming the OpenAI models. In contrast, o3 mini demonstrates the worst performance, primarily because it tends to overthink the task, which is reflected in its insight identifier performance (Section \ref{sec:comp}).

We also report F1 performance of models in the Appendix. Surprisingly, we observe that despite the superior performance of DeepSeek in Exact Match, its performance drops significantly in F1. Upon further investigation, we observe that this is mostly due to DeepSeek's tendency to generate unnecessary content and occasional hallucinations, especially when the right document is not retrieved (we removed the thinking part of DeepSeek-generated answers to calculate the F1). Other models show similar behavior as in Exact Match, with Llama-3.3 70B emerging as the best-performing model. 

Finally, focusing on DeepSeek-R1 because of its superior performance, we report its RAG-based performance when, instead of retrieving documents, we retrieve triples from the set of all extracted triples for each dataset (see the Appendix). We observe that the model shows similar behavior to document-based RAG, but with much less context---since a triple is much shorter than a document---and still falls significantly short compared to Insight-RAG performance. This further highlights the shortcomings of conventional retrieval approaches and the complexity of resolving them.

\subsection{Aggregating Information from Multiple Sources}
\begin{figure*}[th!]
    \centering
    \begin{subfigure}[b]{0.49\textwidth}
        \centering
        \includegraphics[width=\linewidth]{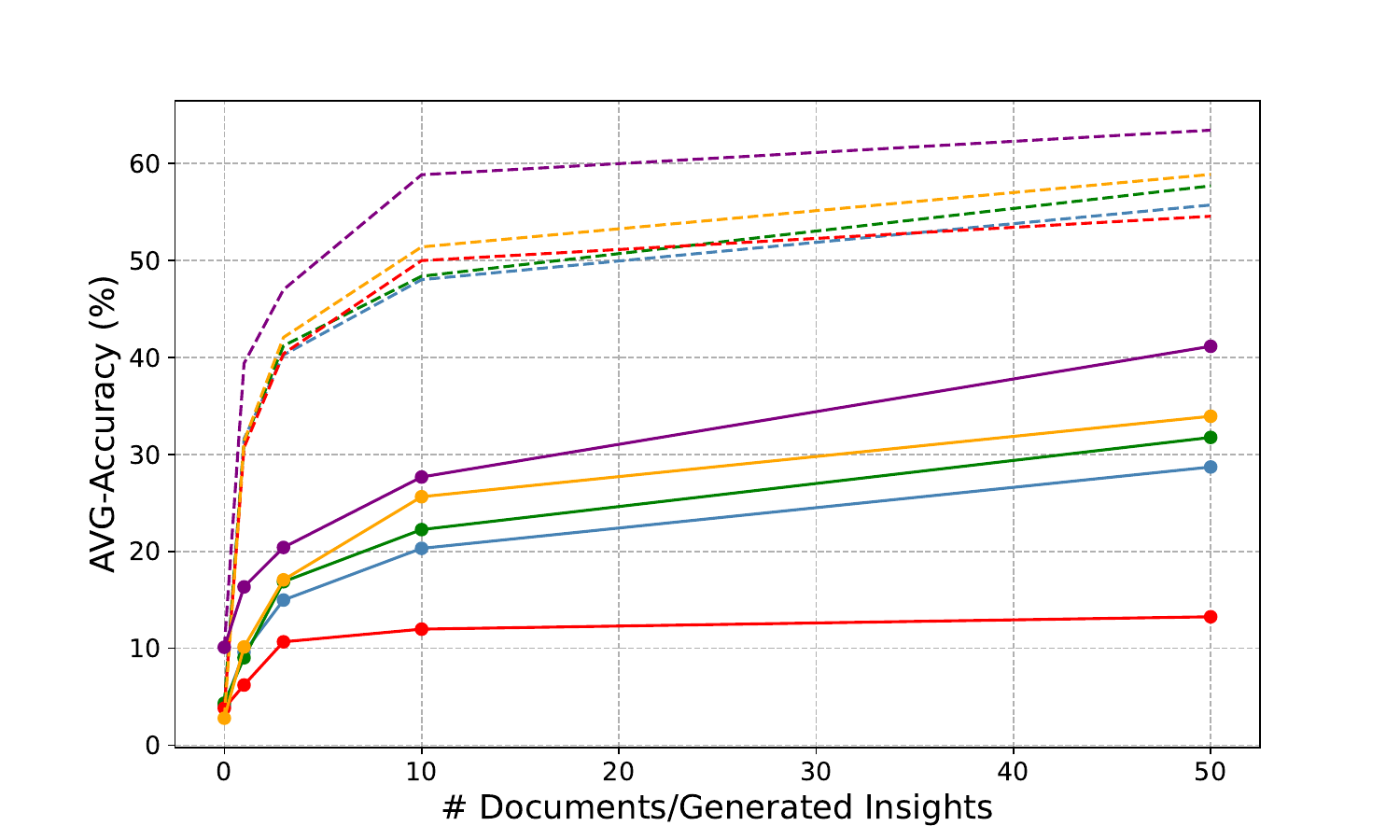}  
        \caption{AAN}
    \end{subfigure}
    \begin{subfigure}[b]{0.49\textwidth}
        \centering
        \includegraphics[width=\linewidth]{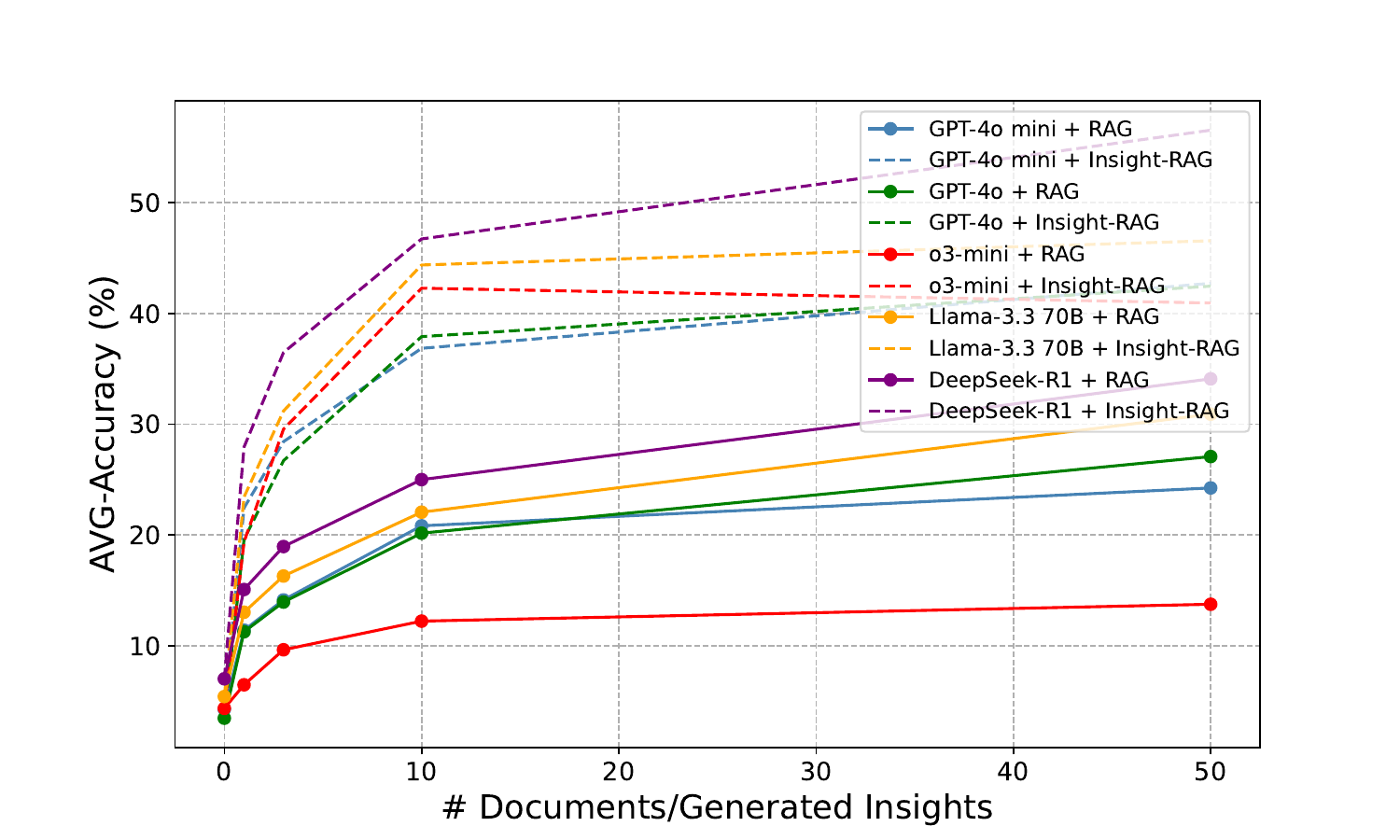}
        \caption{OC}
    \end{subfigure}  
    \caption{The performance comparison of RAG versus Insight-RAG across the AAN and OC datasets in answering questions requiring information from multiple sources. As demonstrated, DeepSeek-R1 performed the best, followed by Llama-3.3 70B. Moreover, we observe that Insight-RAG with only a few generated insights achieves a much higher performance, with the performance continuing to improve at a reduced rate as more insights are added.}
    \label{fig:multi}
\end{figure*}

We present the averaged exact match accuracy (calculated over gold answers for each sample) of Insight-RAG versus conventional RAG using various LLMs for answering questions based on information from multiple sources in Figure \ref{fig:multi}. While using the same number of retrieved documents and generated insights, Insight-RAG consistently outperforms the conventional RAG approach. Moreover, Insight-RAG performance increases rapidly with only a few generated insights, and then its rate of improvement slows down as more generated insights are added. 
Although increasing the number of retrieved documents improves RAG's performance, it still falls short of Insight-RAG, even though the performance gap narrows.  
Overall performance in the multi-source scenario is lower compared to the deeply buried information evaluation, yet the same pattern of Insight-RAG's superiority is evident. Finally, DeepSeek-R1 remains the top performer, followed by Llama, both of which surpass the OpenAI models. 
We also report the average F1 scores and triple-based RAG performance for DeepSeek-R1 in the Appendix. Notably, the performance trends mirror those observed in the F1 metrics for questions on deeply buried information. For triple-based RAG, we observe a degradation in performance---it yields results similar to document-based RAG but when using similar number of tokens in the context.

\begin{table*}[t!]
\small
\centering
\begin{tabular}{lrrrrrr}
\toprule 
\multirow{2}{*}{\bf Model} & \multicolumn{3}{c}{\bf ANN}&  \multicolumn{3}{c}{\bf OC}\\
\cmidrule(lr){2-4}
\cmidrule(lr){5-7}
&Vanilla& RAG (1 doc)&Insight-RAG&Vanilla& RAG (1 doc)&Insight-RAG \\
\midrule
GPT-4o mini&80.8&81.6 \color{cadmiumgreen}{(+0.8)}&82.8 \color{cadmiumgreen}{(+2.0)}&74.4&70.0 \color{cadmiumred}{(-4.4)}&78.0 \color{cadmiumgreen}{(+3.6)}\\
GPT-4o&84.0&80.4 \color{cadmiumred}{(-3.6)}&84.0 (0.0)&71.6&73.6 \color{cadmiumgreen}{(+2.0)}&74.0 \color{cadmiumgreen}{(+2.4)}\\
o3 mini&85.4&85.6 \color{cadmiumgreen}{(+0.2)}&85.6 \color{cadmiumgreen}{(+0.2)}&77.0&74.2 \color{cadmiumred}{(-2.8)}&82.0 \color{cadmiumgreen}{(+5.0)}\\
Llama 3.3 70B&83.8&79.2 \color{cadmiumred}{(-4.6)}&84.4 \color{cadmiumgreen}{(+0.6)}&79.0&77.8 \color{cadmiumred}{(-1.2)}&81.4 \color{cadmiumgreen}{(+2.4)}\\
DeepSeek-R1&70.4&74.0 \color{cadmiumgreen}{(+3.6)}&73.8 \color{cadmiumgreen}{(+3.4)}&66.6&71.4 \color{cadmiumgreen}{(+4.8)}& 72.0 \color{cadmiumgreen}{(+5.4)}\\
\bottomrule
\end{tabular}
\caption{The performance comparison of RAG versus Insight-RAG across the AAN and OC datasets in the paper matching task. As demonstrated, o3 mini performs the best while DeepSeek-R1 shows the lowest performance. Moreover, we observe that Insight-RAG consistently improves performance across all models, while RAG-based solutions show mixed impacts on model performance.}
\label{tab:match_perf}
\postspace
\end{table*}
\subsection{RAG in Non-QA Tasks}
In this section, we evaluate RAG-based solutions on a non-question answering task—specifically, a matching task for citation recommendation. For the RAG baseline, we retrieve only one document because the matching task is not well-defined for traditional RAG approaches, and our experiments did not show any improvement when retrieving additional documents.

Our results, presented in Table \ref{tab:match_perf}, indicate that Insight-RAG consistently outperforms the conventional RAG baseline. This improvement is more pronounced on the OC dataset, likely due to the lower zero-shot performance of the LLMs on that dataset. 
The subjective nature of the matching task (particularly in the AAN dataset) constrains the potential for improvement, resulting in a modest performance gain. 
Furthermore, the RAG baseline demonstrates mixed impacts—yielding both positive and negative effects on model performance across different configurations. Notably, the o3 mini achieves the best overall performance, whereas DeepSeek-R1 performs the worst. Upon further investigation, we found that DeepSeek-R1 tends to unnecessarily overthink the task, which negatively impacts its performance. These findings underscore the effectiveness of the insight-driven approach in extending RAG to tasks beyond question answering and highlight the need for tailored retrieval strategies in non-QA contexts.

\begin{figure}[t!]
    \centering
    \includegraphics[width=\linewidth]{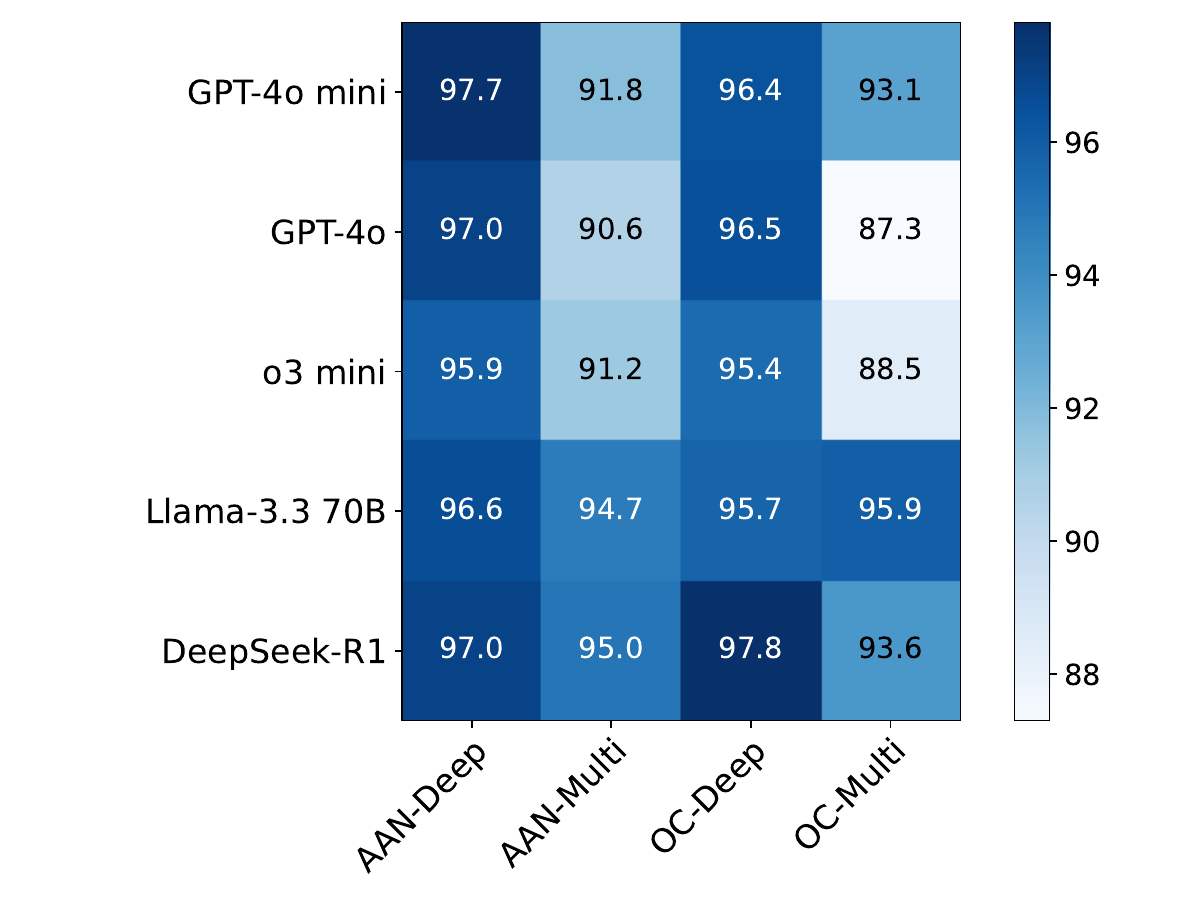}
    \caption{\textbf{Insight Identifier performance:} We ask GPT-4o mini to score the identified insights compared to the gold insights using a three-point scale: 0 (not similar), 0.5 (partially similar), and 1 (completely similar).}
    \label{fig:identifier}
    \postspace
\end{figure}
\subsection{Components Analysis}
\label{sec:comp}
In this section, we analyze the performance of the two key components of the Insight-RAG framework—Insight Identifier and Insight Miner—in addition to the retriever performance of RAG baselines, and discuss how their individual contributions drive the overall success of the systems.

\paragraph{Insight Identifier:}
The Insight Identifier plays a crucial role by processing the input query and distilling the essential informational requirements. To measure the accuracy of the Insight Identifier for deeply buried and multi-source questions, we compare the identified insights with the gold insights (which are concatenations of the subject and relation used to generate the questions). We ask GPT-4o mini to score their similarity using a three-point scale: 0 (not similar), 0.5 (partially similar), and 1 (completely similar). We provide the prompt in the Appendix.

As shown in Figure \ref{fig:identifier}, all models demonstrate high performance in identifying insights given simple questions. The o3 mini shows the lowest performance, which aligns with its overall accuracy in answering the questions. We speculate that this is mostly due to this model's tendency to overthink the task. Moreover, all models show lower performance in multi-source questions compared to deeply buried questions, which is due to the fact that when GPT-4o mini translates triples into question format, it tends to add more unnecessary words in multi-source questions (to capture the fact that there is more than one answer).

\paragraph{Insight Miner:}
We calculate the accuracy of the Insight Miner in predicting the object given the concatenation of subject and relation used to create questions in both deeply buried and multi-source questions. Table \ref{tab:miner} summarizes the Insight Miner's performance based on exact match accuracy for deeply buried questions and recall@10 for multi-source questions, respectively.

Our results indicate that continual pre-training of Llama3.2 3B using LoRA on both the original papers and the extracted triples leads to a reasonably well-performing Insight Miner, with higher performance on deeply buried questions versus multi-source questions. This difference is probably due to the fact that it is easier for the model to learn information about the pair of subject and relation with one object compared to cases when there are multiple objects for a given subject-relation pair.

\paragraph{Retriever:}
Given our knowledge of each question's source paper, we can evaluate the retriever model's accuracy in fetching relevant documents for both deeply buried and multi-source questions. Table \ref{tab:retriever} presents the retriever performance using Hits@50 and MRR metrics, along with their averaged values for multi-source questions. 
As shown, retriever performance is consistently low across all settings, which explains the poor performance of the RAG-based baselines. We attribute this low performance to two primary factors: first, embedding-based representations struggle to capture deeply buried concepts within documents; second, our question generation method produces phrasing that differs from the original text, making it more challenging for the retriever to retrieve the correct document. Additionally, we observe similar performance levels in both deeply buried information and multi-source settings. 

\begin{table}[t!]
\small
\centering
\begin{tabular}{lrr}
\toprule 
\bf Task Type & \bf ANN&  \bf OC\\
\midrule
Deep-Insight&92.1&96.5\\
Multi-Source&72.1&74.8\\
\bottomrule
\end{tabular}
\caption{\textbf{The Insight Miner performance:} We report exact match for deeply buried questions and Recall@10 for multiple source questions.}
\label{tab:miner}
\postspace
\end{table}

\begin{table}[t!]
\small
\centering
\begin{tabular}{lrrrr}
\toprule 
\multirow{2}{*}{\bf Data} & \multicolumn{2}{c}{\bf Deep-Insight}&
\multicolumn{2}{c}{\bf Multi-Source}\\
\cmidrule(lr){2-3}
\cmidrule(lr){4-5}
&Hits@50&MRR&A-Hits@50&A-MRR\\
\midrule
AAN&39.3&0.13&46.8&0.16\\
OC&56.1&0.24&49.5&20.3\\
\bottomrule
\end{tabular}
\caption{\textbf{The retriever performance:} We report Hits@50 and MRR for deeply buried questions and the averaged Hits@50 and MRR for multiple source questions.}
\label{tab:retriever}
\postspace
\end{table}

\begin{figure*}[th!]
    \centering
    \begin{subfigure}[b]{0.31\textwidth}
        \centering
        \includegraphics[width=\linewidth]{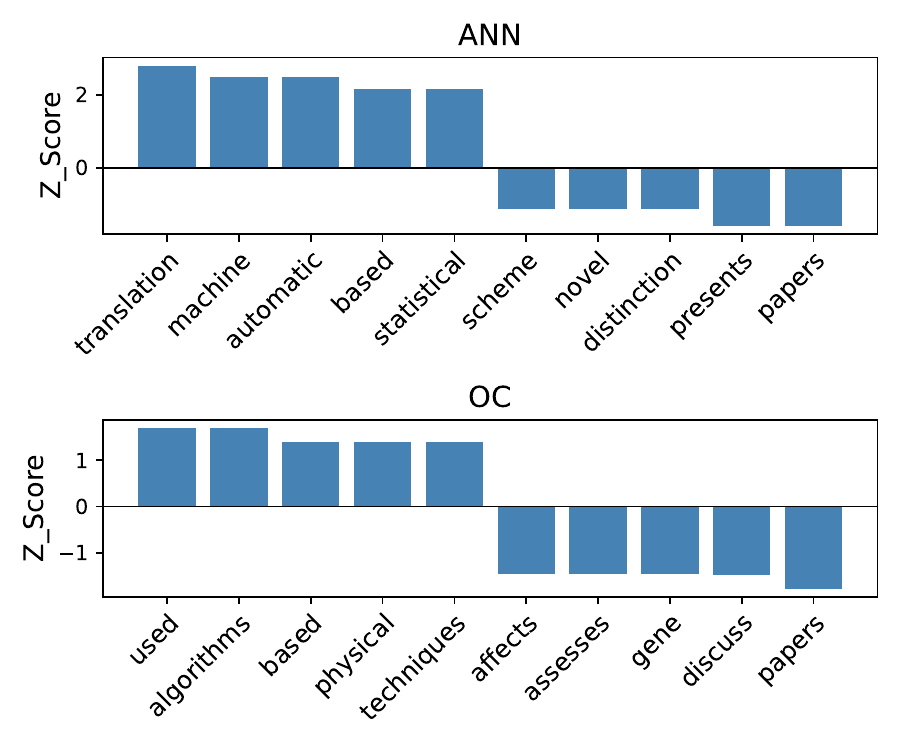}  
        \caption{GPT-4o mini}
    \end{subfigure}
    \begin{subfigure}[b]{0.31\textwidth}
        \centering
        \includegraphics[width=\linewidth]{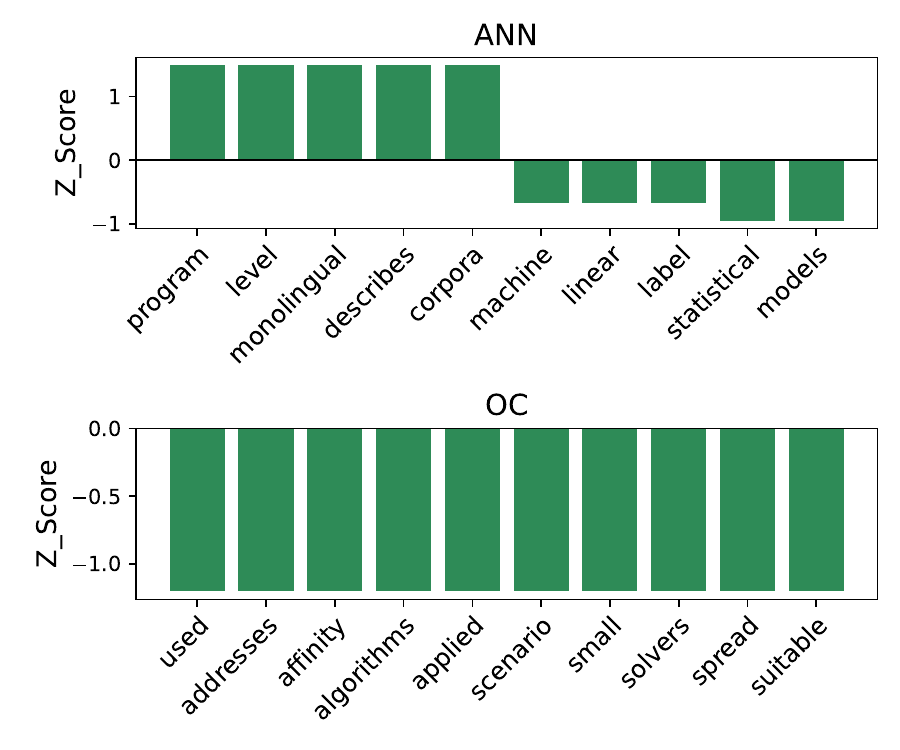}
        \caption{GPT-4o}
    \end{subfigure}  
    \begin{subfigure}[b]{0.31\textwidth}
        \centering
        \includegraphics[width=\linewidth]{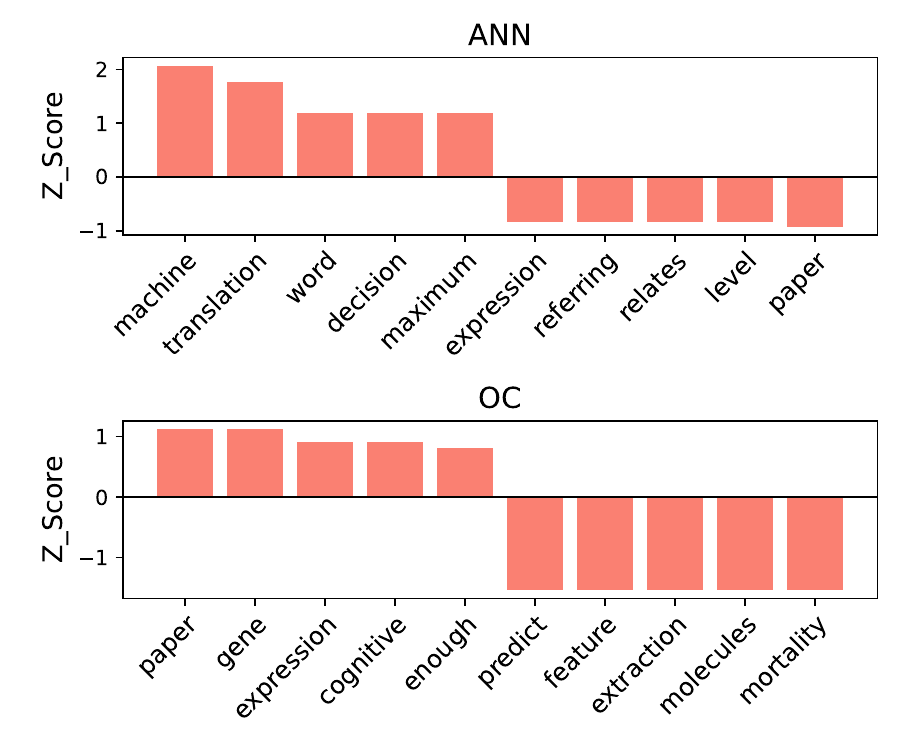}
        \caption{o3 mini}
    \end{subfigure}  
    \begin{subfigure}[b]{0.31\textwidth}
        \centering
        \includegraphics[width=\linewidth]{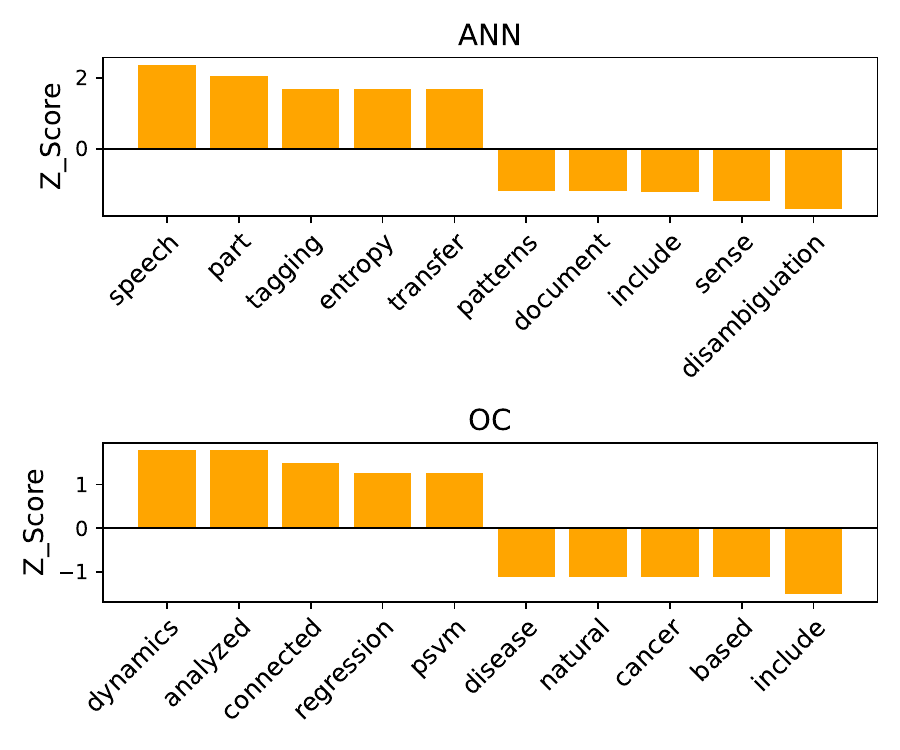}
        \caption{Llama-3.3 70B}
    \end{subfigure}  
    \begin{subfigure}[b]{0.31\textwidth}
        \centering
        \includegraphics[width=\linewidth]{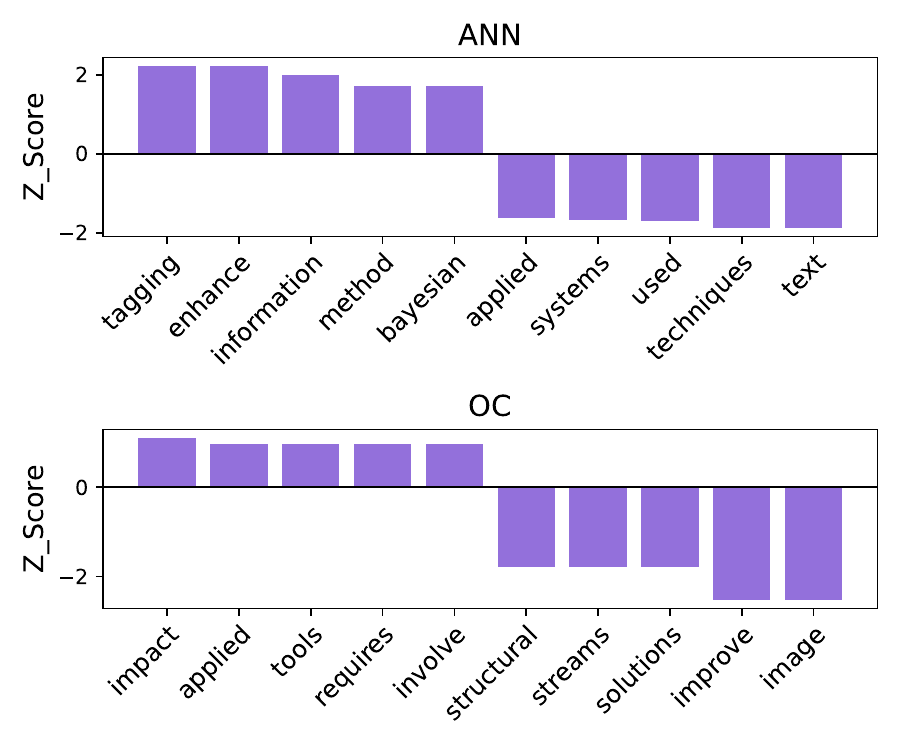}
        \caption{DeepSeek-R1}
    \end{subfigure}   
    \caption{\textbf{The quality of identified insights in the matching task:} We identified the top-5 most positively and negatively influential words in the identified insights using Z-score metrics for each LLM.}
    \label{fig:match_z}
    \postspace
\end{figure*}
\subsection{Identified Insights in Non-QA Tasks}
To better understand the identified insights and their impact on the matching task, we first extract the insights generated by the Insight Identifier module for each model and dataset. We then assign a binary label (0 or 1) to each sample, indicating whether augmenting the sample with these insights changes the model's prediction from correct to incorrect or vice versa, respectively. Next, we identify words with positive or negative impact by calculating the Z-score—a metric introduced to detect artifacts in textual datasets by measuring the correlation between the occurrence of each word and the corresponding sample label \citep{gardner2021competency}.

The Z-score results for the LLMs are shown in Figure \ref{fig:match_z}. Despite the fact that in the prompt we clearly asked the models to identify insights independent of the input identifiers (i.e., Paper A and Paper B), we observe that "paper" appears as an influential token in insights identified by GPT-4o mini and o3 mini, mostly as a negative factor except for o3 mini in the OC dataset.

Overall, OpenAI models appear to benefit from relation words that indicate direct application or description  (e.g., ``used'', ``based'', and ``describes''), while they are hindered by more discursive or predictive terms (e.g., ``presents'', ``discuss'', ``relates'', and ``predict''). In contrast, open LLMs perform better when relations emphasize analytical or connective processes (e.g., ``analyzed'', ``connected'', ``enhance'', and ``involve''), with generic or usage-based terms impairing their performance (e.g., ``include'', ``based'', ``used'', and ``applied''). This indicates that the same relation word can affect different models in opposite ways, highlighting the significant role of model architecture and training history in interpreting relational cues.
Finally, we observe that for GPT-4o, most identified insights did not result in changes to model predictions, suggesting that the Z-scores for this model may not be very trustworthy.

\section{Related Works}
RAG has emerged as a prominent strategy for enhancing LLMs by grounding their responses in external document repositories. Early works in this area focused on integrating retrieval mechanisms with LLMs to improve accuracy and contextual relevance in tasks such as open-domain question answering and summarization \citep{lewis2020retrieval, karpukhin2020dense, guu2020retrieval}. However, conventional RAG approaches typically rely on surface-level matching techniques, which can miss deeper, context-specific information and fail to capture nuanced insights embedded within texts.
More complex versions of RAG, such as Iter-RetGen \citep{shao2023enhancing} and self-RAG \citep{asai2023self}, have also been proposed to handle decomposable and multi-step reasoning tasks \citep{zhao2024retrieval}. However, these methods were not applicable to our experiments since we focus on atomic questions that are not decomposable. Notably, these sophisticated approaches could potentially be integrated on top of Insight-RAG to further enhance performance in scenarios requiring iterative refinement.

Parallel to these developments, research on insight extraction has demonstrated the value of identifying critical, often overlooked details within documents. For example, transformer-based approaches such as OpenIE6 \citep{kolluru2020openie6} have advanced Open Information Extraction by leveraging pretraining to capture nuanced relational data from unstructured text. LLMs have emerged as powerful tools for keyphrase extraction \citep{muhammad2024pre}, and in recent years, they have been increasingly adopted to mine insights from documents across various domains \citep{ma2023demonstration, zhang2023insight, schilling2024text}.

\section{Conclusion and Future Work}
We introduced Insight-RAG, a novel framework that enhances traditional RAG by incorporating an intermediary insight extraction process. Our approach specifically addresses key challenges in conventional RAG pipelines—namely, capturing deeply buried information, aggregating multi-source insights, and extending beyond standard question answering tasks. By evaluating Insight-RAG on our developed targeted benchmarks across two scientific paper datasets (AAN and OC), we have demonstrated that leveraging insight-driven retrieval consistently improves performance. 
Moreover, through detailed component analysis, we identified both the reasoning behind Insight-RAG's superior performance and the factors contributing to the traditional RAG pipeline's poor performance.

Looking forward, Insight-RAG offers promising directions for future research in various ways: (1) Moving beyond the citation recommendation and matching tasks explored in this study, the framework can be extended to various domains, including legal analysis, medical research, business intelligence, and creative content generation. (2) Future work could develop hierarchical insight extraction methods that categorize insights by importance, abstraction level, and relevance, enabling more nuanced retrieval strategies. (3) Extending the framework to handle multimodal data would allow for insight extraction from images, audio, and video alongside text, creating a more comprehensive understanding of complex information ecosystems. (4) Incorporating expert feedback loops would allow domain specialists to guide the insight extraction process, particularly in highly specialized fields where nuanced understanding is critical. (5) Investigating the transferability of insights across domains could reduce the need for domain-specific training while maintaining high performance.

\section{Limitations}
While Insight-RAG offers significant improvements over conventional RAG methods, several limitations must be acknowledged. First, to capture new knowledge and remain current with evolving information, the Insight Miner requires periodic re-training—a process that conventional RAG systems can avoid by directly retrieving documents from an up-to-date corpus. This re-training requirement increases both maintenance complexity and computational overhead. More details are provided in the Appendix.

Additionally, the multi-stage design of Insight-RAG introduces increased computational complexity and potential latency, which may hinder its applicability in real-time or resource-constrained environments. The framework's reliance on carefully crafted prompts for the Insight Identifier also represents a limitation; minor deviations in prompt design can lead to inconsistencies in the extraction of critical insights, affecting downstream performance.

Error propagation across the pipeline is another concern. Inaccuracies in insight identification may lead to misdirected retrieval efforts, ultimately impacting the overall quality of the generated response. Finally, our evaluation has been primarily conducted on scientific paper datasets, which raises questions about the generalizability of the approach to other domains or more unstructured data sources. Future work should explore broader applications and optimize the framework to address these challenges.

\bibliography{custom}

\appendix

\section{Prompts}
The prompts used for the Insight Identifier, question answering with and without augmentation, matching with and without augmentation, and evaluating the identified insights are provided in prompts \ref{prompt:identify}, \ref{prompt:QA}, \ref{prompt:a-QA}, \ref{prompt:matching}, \ref{prompt:a-matching}, and \ref{prompt:insight-eval}, respectively.

\begin{prompt}[title={\footnotesize\texttt{Insight Identifier}}, label=prompt:identify]
You are given a question or task along with its required input. Your goal is to extract the necessary insight that will allow another autoregressive LLM—pretrained on a dataset of scientific papers—to complete the answer. The insight must be expressed as a sentence fragment (i.e., a sentence that is meant to be completed).\\
\\
Instructions:\\
\\
Extract the Insight:\\
Identify the key information needed from the dataset to solve the task or answer the question.\\
Format the insight as a sentence fragment that can be completed by the LLM trained on the dataset.\\
For example, if the task is to find the birthplace of Person X, your insight should be:
"Person X was born in".
\\\\
Determine Answer Multiplicity:\\
Determine whether the answer should be singular or plural based solely on the plurality of the nouns in the question. Do not use common sense or external context—rely exclusively on grammatical cues in the question. \\
For instance, if the question uses plural nouns (e.g., "What are the cities in California?"), set Multi-answer to True. Conversely, if the question uses singular nouns (e.g., "What does pizza contain?"), set it to False.
\\\\
Relevance Check:\\
Only include insights that are directly answerable from the dataset.\\
If an insight does not relate to the available dataset, ignore it.\\
\\
Output Format:\\
Return the result as a list of dictionaries.\\
Each dictionary must have two keys:\\
"Insight": The sentence fragment containing the key insight.\\
"Multi-answer": A Boolean (True or False) indicating whether multiple answers are required.\\
Example Output for follwing questions, Where was Person X born in? what does pizza contain? What are the Cities in California?:\\
\\
{[\\
  \{"Insight": "Person X was born in", "Multi-answer": false\},\\
  \{"Insight": "Pizza contains", "Multi-answer": false\},\\
  \{"Insight": "The cities in California are", "Multi-answer": true\}\\
]}\\
\\
Please provide your final answer in this JSON-like list-of-dictionaries format with no additional commentary.\\
Also, make sure to NOT add any extra word to the insights other than the word present in the input.\\
Remove all unnecessary words and provide the insight in its simplest form. For example, if the query asks "what are the components that X uses?", the insight should be "X uses". Similarly, if the query asks "what are all the components/techniques/features/applications included in Z?", the insight should be "Z include".\\
If a non-question task is given, possible insights might involve asking about how two concepts are connected or a definition of a concept. Only identify the insight you believe will help solve the task, and provide it as a short sentence fragment to be completed. Do not add any unnecessary content or summaries of the input.\\
Additionally, for non-question tasks, the insight should NOT refer to the specific input or include any input-specific identifiers. Instead, it should be a STAND-ALONE statement focusing on the underlying concepts, entities, and their relationships from the inputs. If you cannot find any such insights, return a list of EMPTY dictionary.\\

Task:\\
\{\}
\end{prompt}
\begin{prompt}[title={\footnotesize\texttt{QA}}, label=prompt:QA]
Answer the question. Do not include any extra explanation.\\
Question: \{\}\\
\end{prompt}
\begin{prompt}[title={\footnotesize\texttt{Augmented QA}}, label=prompt:a-QA]
Answer the question using the context. Do not include any extra explanation.\\
Question: \{\}\\
Context: \{\}
\end{prompt}

\begin{prompt}[title={\footnotesize\texttt{Matching}}, label=prompt:matching]
You are provided with two research papers, Paper-A and Paper-B. Your task is to determine if the papers are relevant enough to be cited by the other.\
Your response must be provided in a JSON format with two keys:\\
"explanation": A detailed explanation of your reasoning and analysis.\\
"answer": The final determination ("Yes" or "No").\\
\\
Paper-A:\\
\{\}\\\\

Paper-B:\\
\{\}
\end{prompt}
\begin{prompt}[title={\footnotesize\texttt{Augmented Matching}}, label=prompt:a-matching]
You are provided with two research papers, Paper-A and Paper-B, and some useful insights. Your task is to determine if the papers are relevant enough to be cited by the other. You may use the insights to better predict whether the papers are relevant or not. The insights should only serve as supportive evidence; do not rely on them blindly.\\
Your response must be provided in a JSON format with two keys:\\
"explanation": A detailed explanation of your reasoning and analysis.\\
"answer": The final determination ("Yes" or "No").\\
\\
Paper-A:\\
\{\}\\\\

Paper-B:\\
\{\}
\\\\
Useful insights:\\
\{\}
\end{prompt}

\begin{prompt}[title={\footnotesize\texttt{Identified Insights Evaluation}}, label=prompt:insight-eval]
You are given two incomplete sentences: a target sentence and a generated sentence. Your task is to evaluate how similar these two incomplete sentences are in terms of meaning and content. Please follow these instructions:\\
\\
Similarity Criteria:\\
\\
0: The sentences are not similar at all.\\
0.5: The sentences share some elements or meaning, but are only partially similar.\\
1: The sentences are very similar or essentially equivalent in meaning.\\
\\
Output Requirement:\\
\\
Provide only the similarity score (0, 0.5, or 1) as your output.\\
Do not include any additional text or explanation.
The output format should be as follownig:\\
\\
Score: <0, 0.5, or 1)>
\\\\
Target Sentence: \{\}\\
Generated Sentence: \{\}\\
\end{prompt}

\section{Experimental Details}
\paragraph{Benchmarking:}
We use the processed abstracts from the AAN dataset \citep{radev2013acl} and the OC dataset \citep{bhagavatula2018content}, as provided by \citet{zhou2020multilevel}. This curated set includes approximately 13,000 paper abstracts from AAN and 567,000 abstracts from OC, offering a rich and diverse corpus of academic content. 
Specifically, the AAN dataset comprises computational linguistics papers published in the ACL Anthology from 2001 to 2014, along with their associated metadata, while the OC dataset encompasses approximately 7.1 million papers covering topics in computer science and neuroscience.

\begin{figure*}[th!]
    \centering
    \begin{subfigure}[b]{0.49\textwidth}
        \centering
        \includegraphics[width=\linewidth]{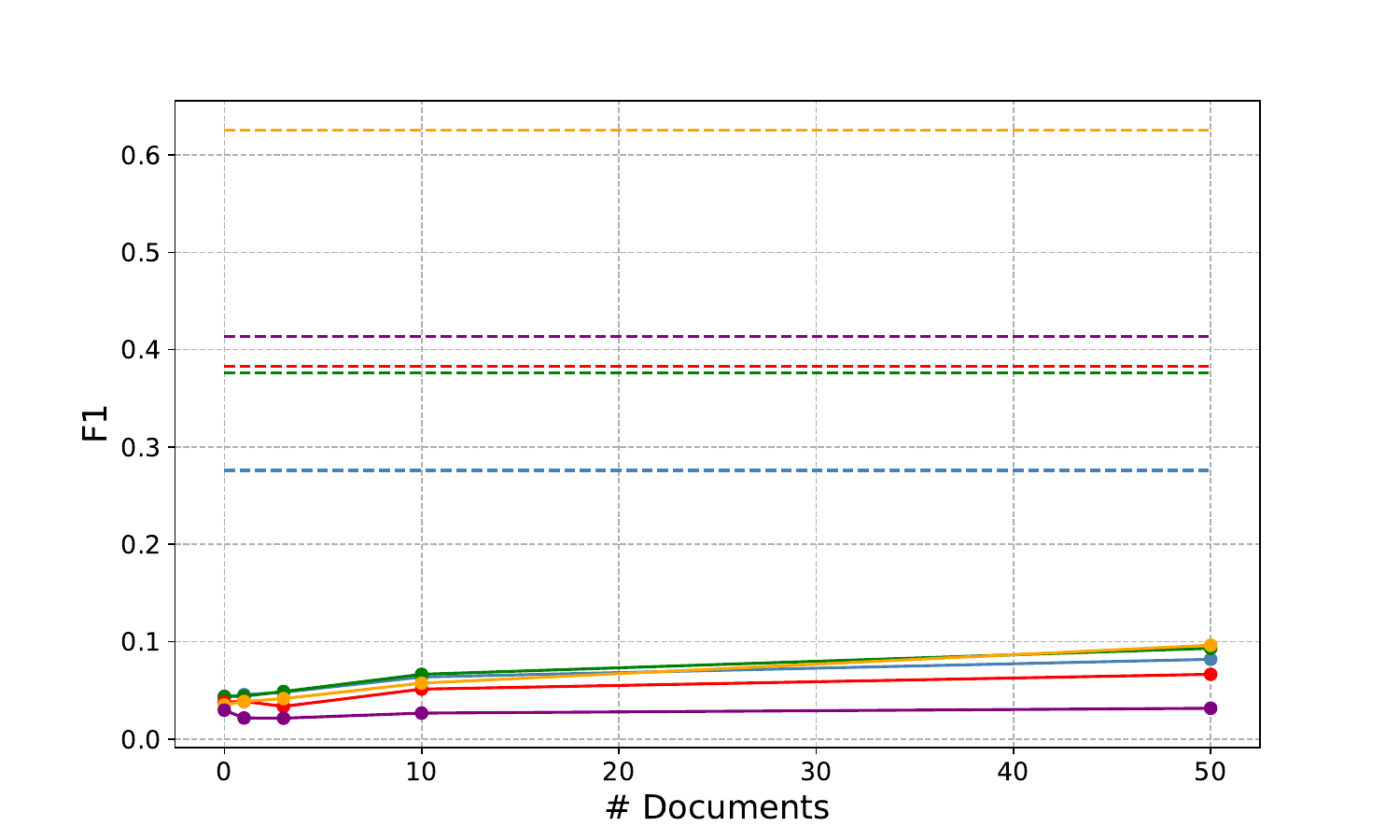}  
        \caption{AAN}
    \end{subfigure}
    \begin{subfigure}[b]{0.49\textwidth}
        \centering
        \includegraphics[width=\linewidth]{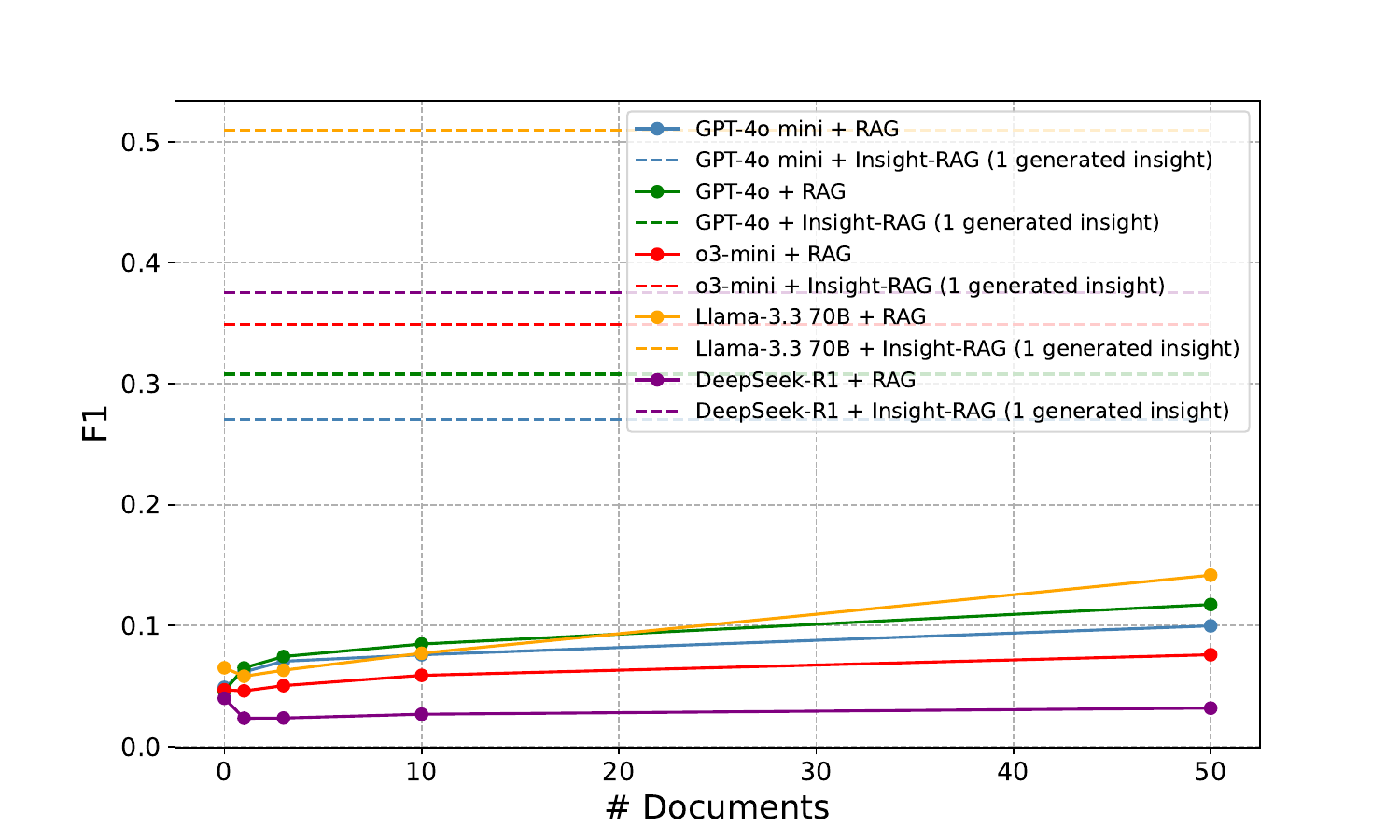}
        \caption{OC}
    \end{subfigure}  
    \caption{The performance comparison of RAG versus Insight-RAG across the AAN and OC datasets based on F1 metric for deeply buried information. As demonstrated, Llama-3.3 performed the best, while DeepSeek-R1 performed the worst.}
    \label{fig:deep-f1}
\end{figure*}

\begin{figure*}[th!]
    \centering
    \begin{subfigure}[b]{0.49\textwidth}
        \centering
        \includegraphics[width=\linewidth]{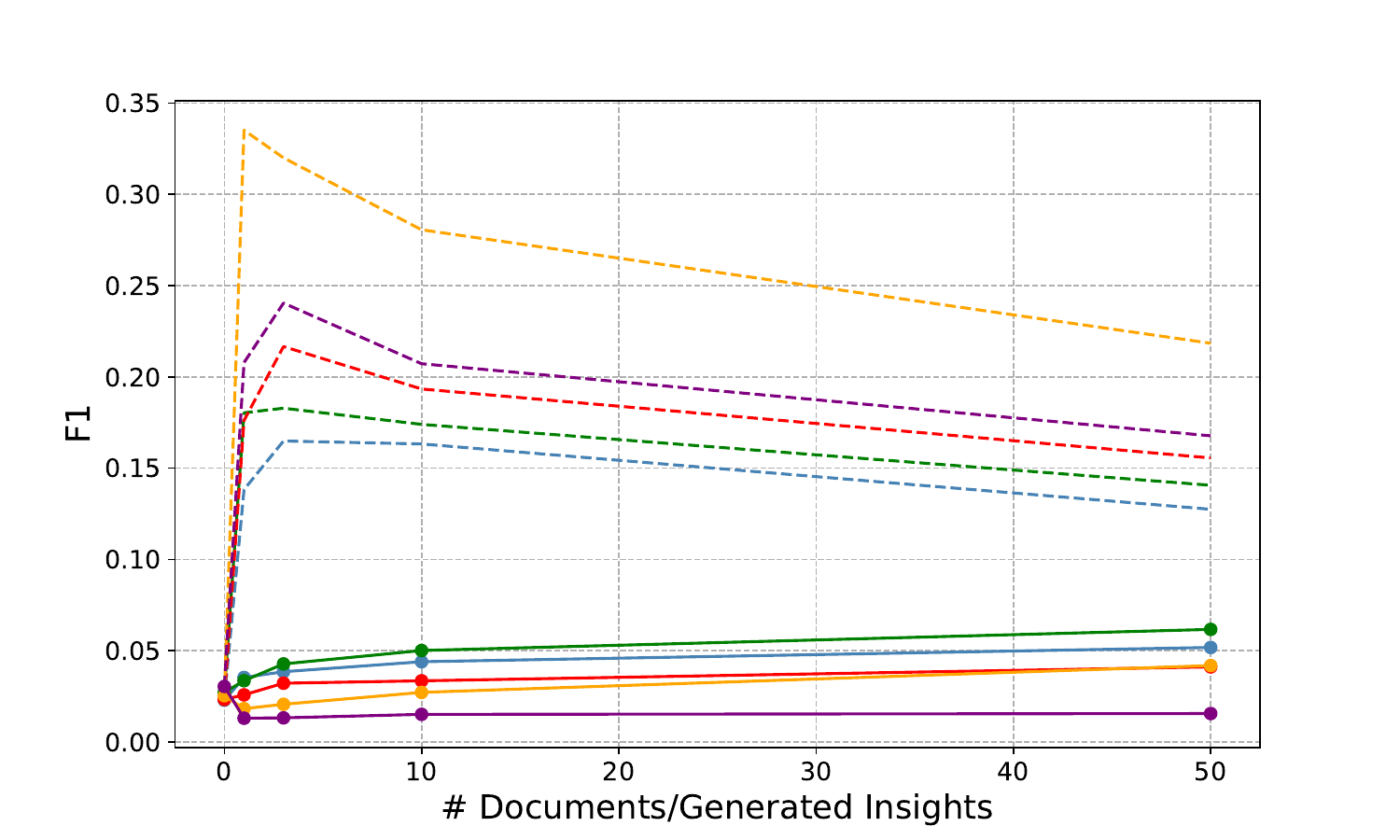}  
        \caption{AAN}
    \end{subfigure}
    \begin{subfigure}[b]{0.49\textwidth}
        \centering
        \includegraphics[width=\linewidth]{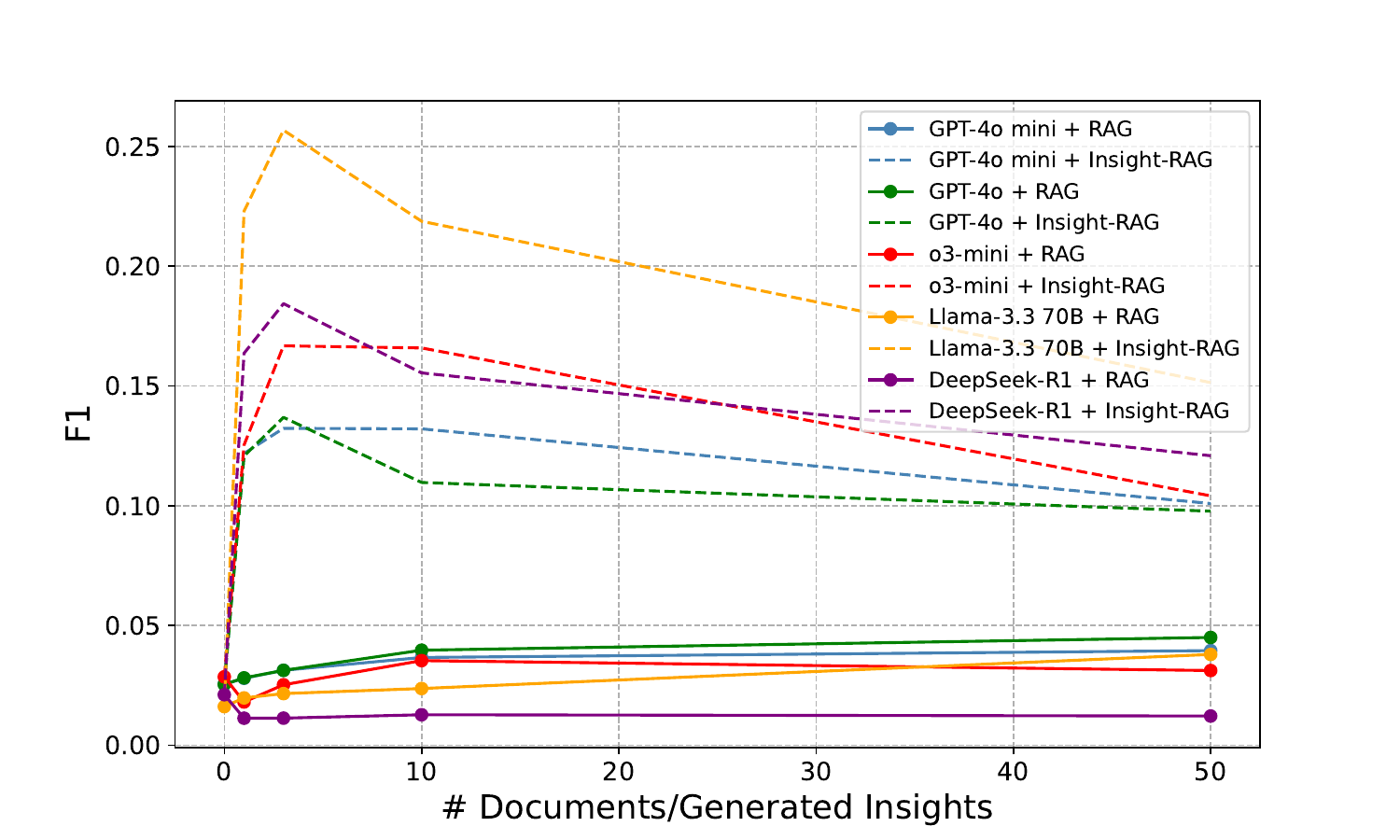}
        \caption{OC}
    \end{subfigure}  
    \caption{The performance comparison of RAG versus Insight-RAG across the AAN and OC datasets based on averaged F1 metric for multi-source questions. As demonstrated, Llama-3.3 performed the best, while DeepSeek-R1 performed the worst.}
    \label{fig:multi-f1}
\end{figure*}

\paragraph{Modeling:} 
For Insight Miner, we perform continual pre-training on LLaMA-3.2 3B with LoRA and optimize hyperparameters through grid search based on training loss.
Specifically, following \citet{pezeshkpour2025learning}, we tuned learning rate $\alpha=[3\times10^{-3}, 10^{-3}, 3\times10^{-4}, 10^{-4}, 3\times10^{-5}, 10^{-5}]$; the LoRA rank $r=[4,8,16]$; the LoRA-alpha $\in \{8,16,32\}$; and the LoRA-dropout $\in \{0.05, 0.1\}$. We trained the LLaMA model for 30 epochs.

\paragraph{Cost and Complexity Considerations:} Continual pre-training of the Insight Miner using LoRA on 8 NVIDIA A100 SXM GPUs for 30 epochs per dataset takes approximately 7 hours. 
Regarding prompting costs, although Insight-RAG includes an additional Insight Identifier component compared to conventional RAG, its ability to achieve much higher performance with a much shorter context length results in lower API costs overall. 
Additionally, while the Insight Miner unit requires periodic retraining to incorporate new information, in many settings this update can be performed infrequently. For environments where new information arrives regularly, an online learning-based solution \citep{hoi2021online,liang2024online} can be adopted to update the model incrementally without necessitating a full retraining cycle.

\begin{table*}[t!]
\small
\centering
\begin{tabular}{lrrrrrr}
\toprule 
\multirow{2}{*}{\bf Model} & \multicolumn{3}{c}{\bf ANN}&  \multicolumn{3}{c}{\bf OC}\\
\cmidrule(lr){2-4}
\cmidrule(lr){5-7}
&Vanilla& RAG (1 doc)&Insight-RAG&Vanilla& RAG (1 doc)&Insight-RAG \\
\midrule
GPT-4o mini&78.8&79.9 \color{cadmiumgreen}{(+1.1)}&82.2 \color{cadmiumgreen}{(+3.4)}&66.0&57.9 \color{cadmiumred}{(-8.1)}&72.5 \color{cadmiumgreen}{(+6.5)}\\
GPT-4o&82.4&77.6 \color{cadmiumred}{(-4.8)}&82.8 \color{cadmiumgreen}{(+0.4)}&61.2&66.3 \color{cadmiumgreen}{(+5.1)}&65.6 \color{cadmiumgreen}{(+4.4)}\\
o3 mini&85.0&85.1 \color{cadmiumgreen}{(+0.1)}&85.4 \color{cadmiumgreen}{(+0.4)}&70.4&65.4 \color{cadmiumred}{(-5.0)}&78.9 \color{cadmiumgreen}{(+8.5)}\\
Llama 3.3 70B&83.8&80.0 \color{cadmiumred}{(-3.8)}&84.8 \color{cadmiumgreen}{(+1.0)}&73.8&71.8 \color{cadmiumred}{(-2.0)}&77.8 \color{cadmiumgreen}{(+4.0)}\\
DeepSeek-R1&59.3&66.7 \color{cadmiumgreen}{(+7.4)}&68.6 \color{cadmiumgreen}{(+9.3)}&50.4&60.6 \color{cadmiumgreen}{(+10.2)}& 62.2 \color{cadmiumgreen}{(+11.8)}\\
\bottomrule
\end{tabular}
\caption{The F1 performance comparison of RAG versus Insight-RAG across the AAN and OC datasets in the paper matching task. As demonstrated, o3 mini performs the best while DeepSeek-R1 shows the lowest performance. Moreover, we observe that Insight-RAG consistently improves performance across all models, while RAG-based solutions show mixed impacts on model performance.}
\label{tab:match_perf_f1}
\end{table*}

\begin{table*}[t!]
\small
\centering
\begin{tabular}{lrrrrrrrr}
\toprule 
\multirow{2}{*}{\bf Model} & \multicolumn{4}{c}{\bf ANN}&  \multicolumn{4}{c}{\bf OC}\\
\cmidrule(lr){2-5}
\cmidrule(lr){6-9}
&1 triple& 3 triples&10 triples&50 triples&1 triple& 3 triples&10 triples&50 triples\\
\midrule
DeepSeek-R1 (Deep)&13.8&18.9&25.8&35.2&20.1&27.0&33.0&42.2\\
DeepSeek-R1 (Multi)&12.1&14.0&14.7&25.2&10.6&13.9&17.9&22.7\\
\bottomrule
\end{tabular}
\caption{RAG-based exact match and averaged exact match accuracy of DeepSeek-R1 for deeply buried and multi-source questions. Instead of retrieving documents, we retrieve triples---using the set of extracted triples.}
\label{tab:rag-triple}
\end{table*}

\section{Experimnets}
We report F1 and averaged F1 performance for all models for deeply buried and multi-source questions in Figure \ref{fig:deep-f1} and \ref{fig:multi-f1}, respectively. Interestingly, despite DeepSeek's superior performance in Exact Match metrics, its F1 scores show a significant decline. Upon closer examination, we discovered this discrepancy stems primarily from DeepSeek's tendency to generate excessive content and occasional hallucinations, particularly when the correct document isn't retrieved. This poor F1 performances occur despite our removal of DeepSeek's ``thinking'' sections when calculating F1 scores. The other evaluated models demonstrate performance patterns similar to their Exact Match results, with Llama-3.3 70B consistently emerging as the top-performing model across both setting.  
Moreover, Table \ref{tab:match_perf_f1} presents the F1 scores for the paper matching task. While these results follow similar trends as the accuracy metric, the F1 scores reveal that both the positive and negative impacts of conventional RAG as well as the benefits of Insight-RAG, are even more amplified compared to accuracy.

Finally, focusing on DeepSeek-R1 due to its superior performance, we report its RAG-based results when, instead of retrieving documents, we retrieve triples from the set of all extracted triples for each dataset. Table \ref{tab:rag-triple} provides the exact match accuracy for the deeply buried information setting, along with the averaged exact match accuracy for the multi-source setting. We observe that while the model shows similar behavior to document-based RAG, using much less context---since a triple is much shorter than a document---it still falls significantly short compared to Insight-RAG performance.  The overall gap between triple-based RAG and Insight-RAG underscores the shortcomings of conventional retrieval approaches and the complexity of resolving them.

\end{document}